\documentclass{article}

\usepackage{times}
\usepackage{graphicx} 
\usepackage{subfigure} 

\usepackage{natbib}

\usepackage{algorithm}
\usepackage{algorithmic}

\usepackage{hyperref}

\usepackage{multicol, multirow, amssymb, amsmath}

\input{def.set}

\usepackage[accepted]{icml2015}

\icmltitlerunning{Bitwise Neural Networks}

\begin{document} 

\twocolumn[
\icmltitle{Bitwise Neural Networks}

\icmlauthor{Minje Kim}{minje@illinois.edu}
\icmladdress{Department of Computer Science, University of Illinois at Urbana-Champaign, Urbana, IL 61801 USA}
\icmlauthor{Paris Smaragdis}{paris@illinois.edu}
\icmladdress{University of Illinois at Urbana-Champaign, Urbana, IL 61801 USA\\
Adobe Research, Adobe Systems Inc., San Francisco, CA 94103, USA}

\icmlkeywords{Deep Learning, Neural Networks}

\vskip 0.3in
]

\begin{abstract} 
Based on the assumption that there exists a neural network that efficiently represents a set of Boolean functions between all binary inputs and outputs, we propose a process for developing and deploying neural networks whose weight parameters, bias terms, input, and intermediate hidden layer output signals, are all binary-valued, and require only basic bit logic for the feedforward pass. The proposed Bitwise Neural Network (BNN) is especially suitable for resource-constrained environments, since it replaces either floating or fixed-point arithmetic with significantly more efficient bitwise operations. Hence, the BNN requires for less spatial complexity, less memory bandwidth, and less power consumption in hardware. In order to design such networks, we propose to add a few training schemes, such as weight compression and noisy backpropagation, which result in a bitwise network that performs almost as well as its corresponding real-valued network. We test the proposed network on the MNIST dataset, represented using binary features, and show that BNNs result in competitive performance while offering dramatic computational savings.
\end{abstract}

\section{Introduction}

According to the universal approximation theorem, a single hidden layer with a finite number of units can approximate a continuous function with some mild assumptions \cite{CybenkoG1989mcss, HornikK1991}. While this theorem implies a shallow network with a potentially intractable number of hidden units when it comes to modeling a complicated function, Deep Neural Networks (DNN) achieve the goal by learning a hierarchy of features in their multiple layers \cite{HintonG2006, BengioY2009ftml}.

Although DNNs are extending the state of the art results for various tasks, such as image classification \cite{GoodfellowI2013icml}, speech recognition \cite{HintonG2012ieeespm}, speech enhancement \cite{XuY2014ieeespl}, etc, it is also the case that the relatively bigger networks with more parameters than before call for more resources (processing power, memory, battery time, etc), which are sometimes critically constrained in applications running on embedded devices. Examples of those applications span from context-aware computing, collecting and analysing a variety of sensor signals on the device \cite{BaldaufM2007jahuc}, to always-on computer vision applications (e.g. Google glasses), to speech-driven personal assistant services, such as ``Hey, Siri." A primary concern that hinders those applications from being more successful is that they assume an always-on pattern recognition engine on the device, which will drain the battery fast unless it is carefully implemented to minimize the use of resources. Additionally, even in an environment with the necessary resources being available, speeding up a DNN can greatly improve the user experience when it comes to tasks like searching big databases \cite{SalakhutdinovR2009ijar}. In either case, a more compact yet still well-performing DNN is a welcome improvement.

Efficient computational structures for deploying artificial neural networks have long been studied in the literature. Most of the effort is focused on training networks whose weights can be transformed into some quantized representations with a minimal loss of performance \cite{FieslerE1990isop, HwangK2014sips}. They typically use the quantized weights in the feedforward step at every training iteration, so that the trained weights are robust to the known quantization noise caused by a limited precision. It was also shown that 10 bits and 12 bits are enough to represent gradients and storing weights for implementing the state-of-the-art maxout networks even for training the network \cite{CourbariauxM2014arxiv}. However, in those quantized networks one still needs to employ arithmetic operations, such as multiplication and addition, on fixed-point values. Even though faster than floating point, they still require relatively complex logic and can consume a lot of power.

With the proposed Bitwise Neural Networks (BNN), we take a more extreme view that every input node, output node, and weight, is represented by a single bit. For example, a weight matrix between two hidden layers of 1024 units is a $1024\times 1025$ matrix of binary values rather than quantized real values (including the bias). Although learning those bitwise weights as a Boolean concept is an NP-complete problem \cite{PittL1988jacm}, the bitwise networks have been studied in the limited setting, such as $\mu$-perceptron networks where an input node is allowed to be connected to one and only one hidden node and its final layer is a union of those hidden nodes \cite{GoleaM1992nips}. A more practical network was proposed in \cite{SoudryD2014nips} recently, where the posterior probabilities of the binary weights were sought using the Expectation Back Propagation (EBP) scheme, which is similar to backpropagation in its form, but has some advantages, such as parameter-free learning and a straightforward discretization of the weights. Its promising results on binary text classification tasks however, rely on the real-valued bias terms and averaging of predictions from differently sampled parameters.

This paper presents a completely bitwise network where all participating variables are bipolar binaries. Therefore, in its feedforward only XNOR and bit counting operations are used instead of multiplication, addition, and a nonlinear activation on floating or fixed-point variables. For training, we propose a two-stage approach, whose first part is typical network training with a weight compression technique that helps the real-valued model to easily be converted into a BNN. To train the actual BNN, we use those compressed weights to initialize the BNN parameters, and do noisy backpropagation based on the tentative bitwise parameters. To binarize the input signals, we can adapt any binarization techniques, e.g. fixed-point representations and hash codes. Regardless of the binarization scheme, each input node is given only a single bit at a time, as opposed to a bit packet representing a fixed-point number. This is significantly different from the networks with quantized inputs, where a real-valued signal is quantized into a set of bits, and then all those bits are fed to an input node in place of their corresponding single real value. Lastly, we apply the sign function as our activation function instead of a sigmoid to make sure the input to the next layer is bipolar binary as well. We compare the performance of the proposed BNN with its corresponding ordinary real-valued networks on hand-written digit recognition tasks, and show that the bitwise operations can do the job with a very small performance loss, while providing a large margin of improvement in terms of the necessary computational resources.

\section{Feedforward in Bitwise Neural Networks}
\begin{figure}[ht]
\vskip 0.2in
\begin{center}
\subfigure[]{\includegraphics[scale=.43]{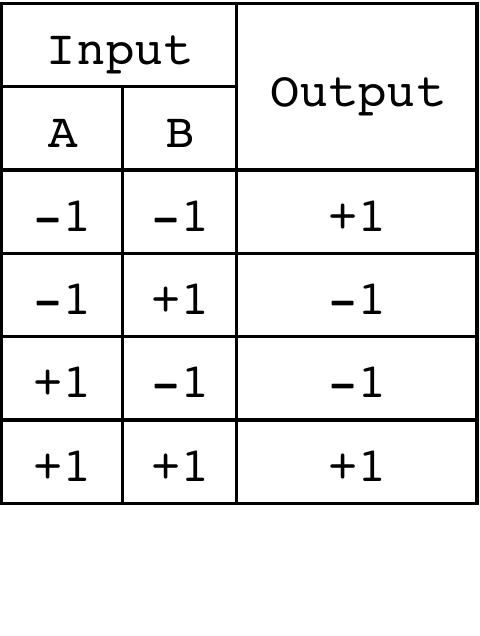}}
\subfigure[]{\includegraphics[scale=.43]{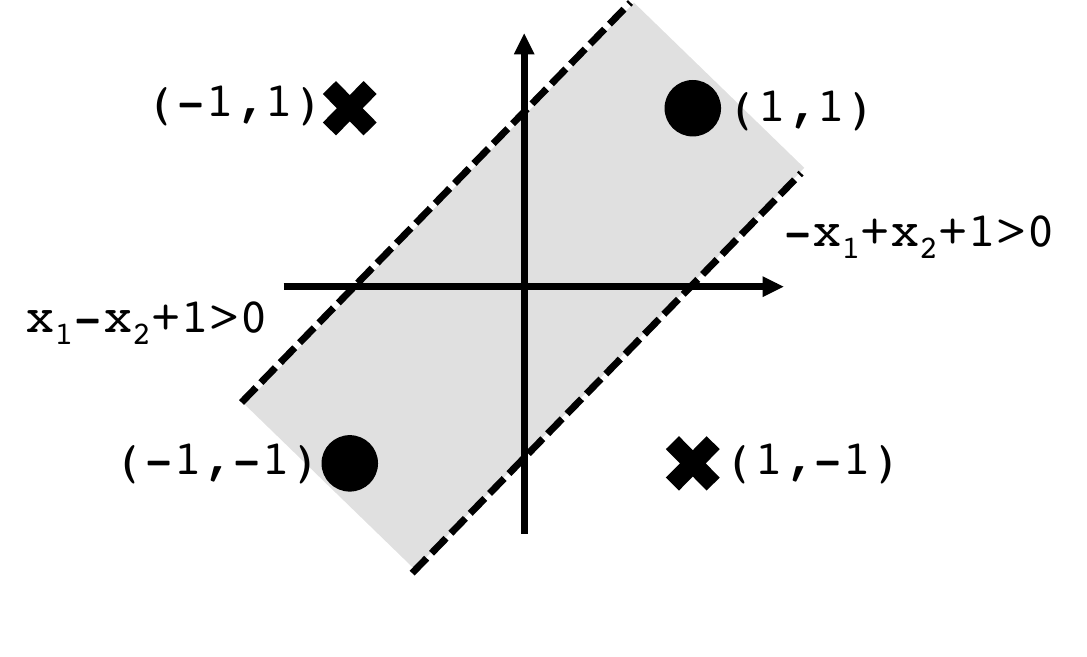}}
\subfigure[]{\includegraphics[scale=.43]{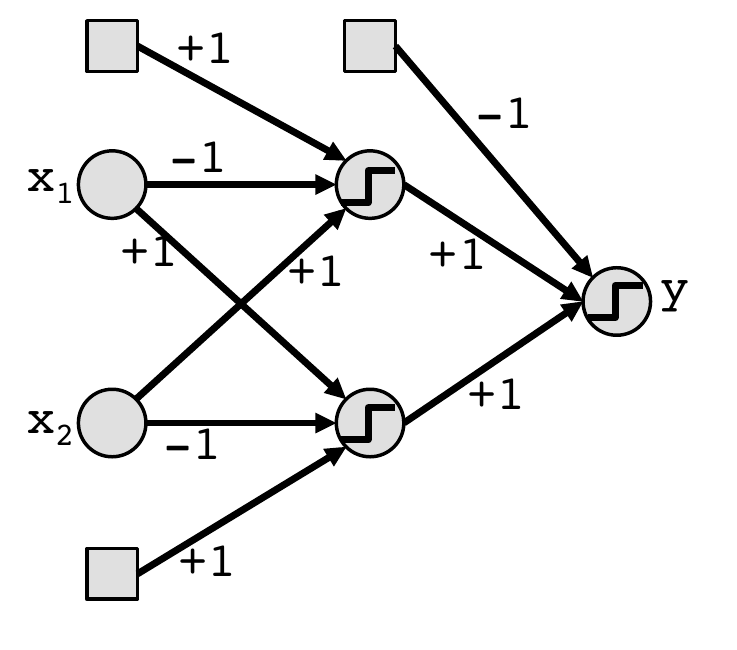}}
\subfigure[]{\includegraphics[scale=.43]{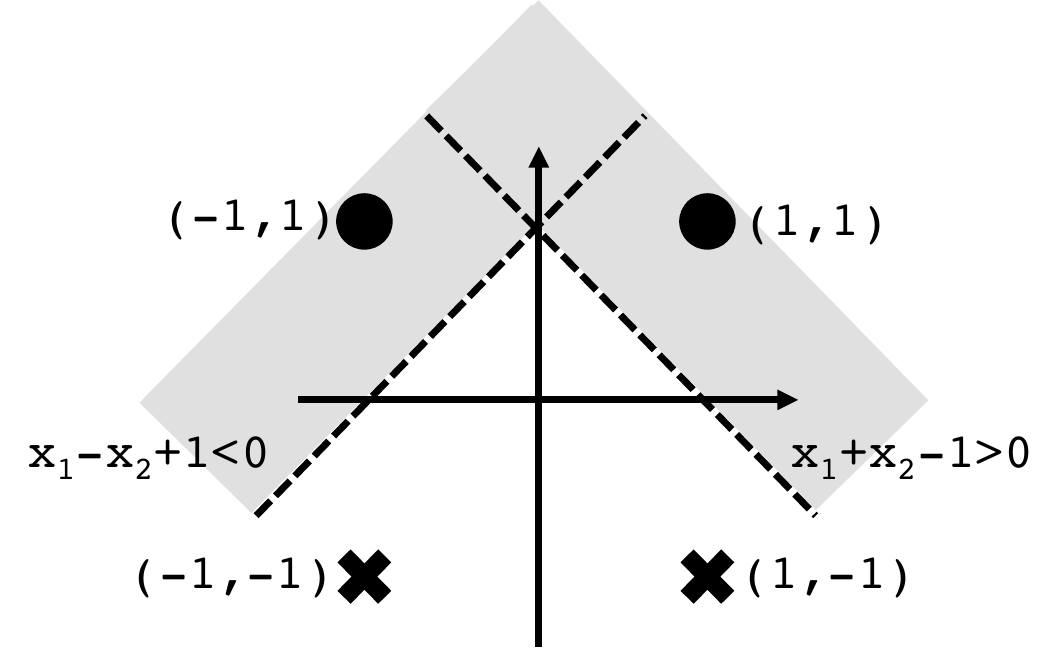}}
\subfigure[]{\includegraphics[scale=.43]{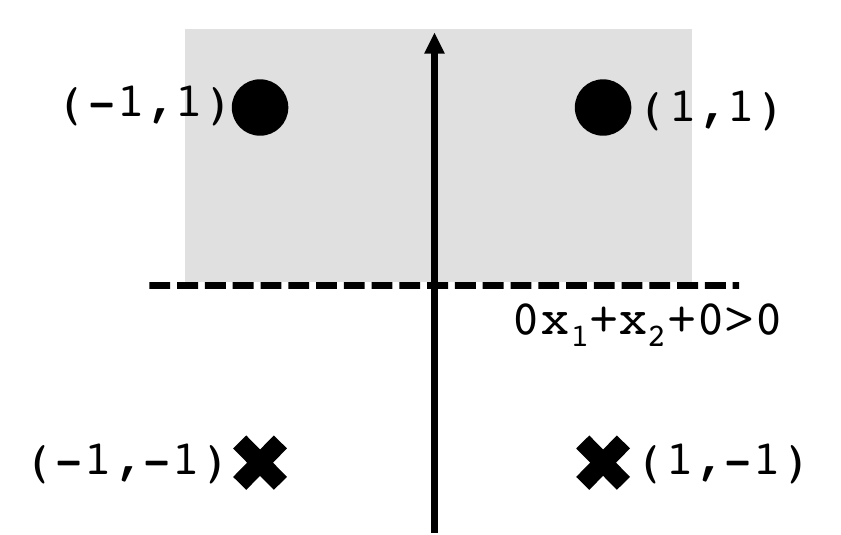}}
\caption{(a) An XNOR table. (b) The XOR problem that needs two hyperplanes. (c) A multi-layer perceptron that solves the XOR problem. (d) A linearly separable problem while bitwise networks need two hyperplanes to solve it ($y=x_2$). (e) A bitwise network with zero weights that solves the $y=x_2$ problem.}
\label{fig:toy}
\end{center}
\vskip -0.2in
\end{figure}

It has long been known that any Boolean function, which takes binary values as input and produces binary outputs as well, can be represented as a bitwise network with one hidden layer \cite{MccullochW1943bmb}, for example, by merely memorizing all the possible mappings between input and output patterns. We define the forward propagation procedure as follows based on the assumption that we have trained such a network with bipolar binary parameters:
\begin{align}
\label{eq:xnor}a_i^{l} &=  b_i^l+\sum_j^{K^{l-1}} w_{i,j}^l \otimes  z_j^{l-1},\\
\label{eq:sign}z_i^{l} &= \textrm{sign}\big(a_i^{l}\big),\\
\bz^l&\in \mathbb{B}^{K^l}, \bW^l\in \mathbb{B}^{K^l\times K^{l-1}}, \bb^l\in \mathbb{B}^{K^l}, 
\end{align}
where $\mathbb{B}$ is the set of bipolar binaries, i.e. $\pm 1$\footnote{In the bipolar binary representation, $+1$ stands for the ``TRUE'' status, while $-1$ is for ``FALSE.''}, and $\otimes$ stands for the bitwise XNOR operation (see Figure \ref{fig:toy} (a)). $l$, $j$, and $i$ indicate a layer, input and output units of the layer, respectively. We use bold characters for a vector (or a matrix if capicalized). $K^l$ is the number of input units at $l$-th layer. Therefore, $\bz^0$ equals to an input vector, where we omit the sample index for the notational convenience. We use the sign activation function to generate the bipolar outputs.

We can check the prediction error $\calE$ by measuring the bitwise agreement of target vector $\bt$ and the output units of $L$-th layer using XNOR as a multiplication operator,
\begin{equation}
\label{eq:err_bits}\calE=\sum_i^{K^{L+1}}\big(1- t_i \otimes  z_i^{L+1}\big)/2,
\end{equation}
but this error function can be tentatively replaced by involving a softmax layer during the training phase.

The XNOR operation is a faster substitute of binary multiplication. Therefore, \eqref{eq:xnor} and \eqref{eq:sign} can be seen as a special version of the ordinary feedforward step that only works when the inputs, weights, and bias are all bipolar binaries. Note that these bipolar bits will in practice be implemented using 0/1 binary values, where \eqref{eq:sign} activation is equivalent to counting the number of 1's and then checking if the accumulation is bigger than the half of the number of input units plus 1. With no loss of generality, in this paper we will use the $\pm 1$ bipolar representation since it is more flexible in defining hyperplanes and examining the network behavior.

Sometimes a BNN can solve the same problem as a real-valued network without any size modifications, but in general we should expect that a BNN could require larger network structures than a real-valued one. For example, the XOR problem in Figure \ref{fig:toy} (b) can have an infinite number of solutions with real-valued parameters once a pair of hyperplanes can successfully discriminate $(1,1)$ and $(-1,-1)$ from $(1,-1)$ and $(-1,1)$. Among all the possible solutions, we can see that binary weights and bias are enough to define the hyperplanes, $x_1-x_2+1>0$ and $-x_1+x_2+1>0$ (dashes). Likewise, the separation performance of the particular BNN defined in (c) has the same classification power once the inputs are binary as well. 

Figure \ref{fig:toy} (d) shows another example where BNN requires more hyperplanes than a real-valued network. This linearly separable problem is solvable with only one hyperplane, such as $-0.1x_1+x_2 +0.5> 0$, but it is impossible to describe such a hyperplane with binary coefficients. We can instead come up with a solution by combining multiple binary hyperplanes that will eventually increase the perceived complexity of the model. However, even with a larger number of nodes, the BNN is not necessarily more complex than the smaller real-valued network. This is because a parameter or a node of BNN  requires only one bit to represent while a real-valued node generally requires more than that, up to 64 bits. Moreover, the simple XNOR and bit counting operations of BNN bypass the computational complications of a real-valued system, such as the power consumption of multipliers and adders for the floating-point operations, various dynamic ranges of the fixed-point representations, erroneous flips of the most significant bits, etc. Note that if the bitwise parameters are sparse, we can further reduce the number of hyperplanes. For example, for an inactive element in the weight matrix $\bW$ due to the sparsity, we can simply ignore the computation for it similarly to the operations on the sparse representations. Conceptually, we can say that those inactive weights serve as zero weights, so that a BNN can solve the problem in Figure \ref{fig:toy} (d) by using only one hyperplane as in (e). From now on, we will use this extended version of BNN with inactive weights, yet there are some cases where BNN needs more hyperplanes than a real-valued network even with the sparsity.
    
\section{Training Bitwise Neural Networks}
We first train some compressed network parameters, and then retrain them using noisy backpropagation for BNNs.

\subsection{Real-valued Networks with Weight Compression}
First, we train a real-valued network that takes either bitwise inputs or real-valued inputs ranged between $-1$ and $+1$. A special part of this network is that we constrain the weights to have values between $-1$ and $+1$ as well by wrapping them with $\tanh$. Similarly, if we choose $\tanh$ for the activation, we can say that the network is a relaxed version of the corresponding bipolar BNN. With this weight compression technique, the relaxed forward pass during training is defined as follows:
\begin{align}
\label{eq:fp_lin}a_i^{l} &=  \tanh(\bar{b}_i^l)+\sum_j^{K^{l-1}} \tanh(\bar{w}_{i,j}^l) \bar{z}_j^{l-1},\\
\label{eq:fp_act}\bar{z}_i^{l} &= \tanh\big(a_i^{l}\big),
\end{align}
where all the binary values in \eqref{eq:xnor} and \eqref{eq:sign} are real for the time being: $\bar{\bW}^l\in\Real^{K^{l}\times K^{l-1}}$, $\bar{\bb}^l\in\Real^{K^{l}}$, and $\bar{\bz}^l\in\Real^{K^l}$. The bars on top of the notations are for the distinction.

Weight compression needs some changes in the backpropagation procedure. In a hidden layer we calculate the error,
\begin{align}
\nonumber\delta_j^{l}(n)&=\Big(\sum_i^{K^{l+1}}\tanh(\bar{w}_{i,j}^{l+1})\delta_i^{l+1}(n)\Big)\cdot\Big(1-\tanh^2\big(a_j^{l}\big)\Big).
\end{align}
Note that the errors fron the next layer are multiplied with the compressed versions of the weights. Hence, the gradients of the parameters in the case of batch learning are
\begin{align}
\nonumber\nabla \bar{w}_{i,j}^{l} &= \Big(\sum_n \delta_i^{l}(n)\bar{z}_j^{l-1}\Big)\cdot\Big(1-\tanh^2\big(\bar{w}_{i,j}^{l}\big)\Big),\\
\nonumber\nabla \bar{b}_{i}^{l} &= \Big(\sum_n\delta_i^{l}(n)\Big)\cdot\Big(1-\tanh^2\big(\bar{b}_{i}^{l}\big)\Big),
\end{align}
with the additional term from the chain rule on the compressed weights.

\subsection{Training BNN with Noisy Backpropagation}\label{sec:bnn_bp}
Since we have trained a real-valued network with a proper range of weights, what we do next is to train the actual bitwise network. The training procedure is similar to the ones with quantized weights \cite{FieslerE1990isop, HwangK2014sips}, except that the values we deal with are all bits, and the operations on them are bitwise. To this end, we first initialize all the real-valued parameters, $\bar{\bW}$ and $\bar{\bb}$, with the ones learned from the previous section. Then, we setup a sparsity parameter $\lambda$ which says the proportion of the zeros after the binarization. Then, we divide the parameters into three groups: $+1$, $0$, or $-1$. Therefore, $\lambda$ decides the boundaries $\beta$, e.g. $w_{ij}^l=-1$ if $\bar{w}_{ij}^l<-\beta$. Note that the number of zero weights $|\bar{w}_{ij}^l|<\beta$ equals to $\lambda K^l K^{l-1}$.

The main idea of this second training phase is to feedforward using the binarized weights and the bit operations as in \eqref{eq:xnor} and \eqref{eq:sign}. Then, during noisy backpropagation the errors and gradients are calculated using those binarized weights and signals as well: 
\begin{align}
\nonumber\delta_j^{l}(n)&=\sum_i^{K^{l+1}}w_{i,j}^{l+1}\delta_i^{l+1}(n),\\
\nabla \bar{w}_{i,j}^{l} &= \sum_n \delta_i^{l}(n)z_j^{l-1}, \quad\nabla \bar{b}_{i}^{l} = \sum_n\delta_i^{l}(n).
\end{align}

In this way, the gradients and errors properly take the binarization of the weights and the signals into account. Since the gradients can get too small to update the binary parameters $\bW$ and $\bb$, we instead update their corresponding real-valued parameters, 
\begin{equation}\label{eq:updates}
\bar{w}_{i,j}^l \leftarrow \bar{w}_{i,j}^l - \eta \nabla \bar{w}_{i,j}^l, \quad  \bar{b}_{i,j}^l \leftarrow \bar{b}_{i}^l - \eta \nabla \bar{b}_{i}^l,
\end{equation}
with $\eta$ as a learning rate parameter. Finally, at the end of each update we binarize them again with $\beta$. We repeat this procedure at every epoch.

\section{Experiments}

In this section we go over the details and results of the hand-written digit recognition task on the MNIST data set \cite{LecunY98procieee} using the proposed BNN system. Throughout the training, we adopt the softmax output layer for these multiclass classification cases. All the networks have three hidden layers with 1024 units per layer.

From the first round of training, we get a regular dropout network with the same setting suggested in \cite{SrivastavaN2014jmlr}, except the fact that we used the hyperbolic tangent for both weight compression and activation to make the network suitable for initializing the following bipolar bitwise network. The number of iterations from $500$ to $1,000$ was enough to build a baseline. The first row of Table \ref{tab:mnist} shows the performance of the baseline real-valued network with 64bits floating-point. As for the input to the real-valued networks, we rescale the pixel intensities into the bipolar range, i.e. from $-1$ to $+1$, for the bipolar case (the first column). In the second column, we use the original input between 0 and 1 as it is. For the third column, we encode the four equally spaced regions between 0 to 1 into two bits, and feed each bit into each input node. Hence, the baseline network for the third input type has $1,568$ binary input nodes rather than $784$ as in the other cases.

\begin{table}[t]
\caption{Classification errors for real-valued and bitwise networks on different types of bitwise features}
\label{tab:mnist}
\vskip 0.15in
\begin{center}
\begin{small}
\begin{sc}
\begin{tabular}{lcccr}
\hline
\abovespace
\multirow{2}{*}{Networks} & \multirow{2}{*}{Bipolar} & \multirow{2}{*}{0 or 1} & Fixed-point \\
\belowspace
 & & & (2bits)\\
\hline
\abovespace
Floating-point & \multirow{2}{*}{1.17\%} & \multirow{2}{*}{1.32\%}& \multirow{2}{*}{1.36\%} \\
\belowspace
networks (64bits)&  &  &  \\
\hline
\abovespace
\belowspace
BNN   & 1.33\%& 1.36\% & 1.47\% \\
\hline
\end{tabular}
\end{sc}
\end{small}
\end{center}
\vskip -0.1in
\end{table}

Once we learn the real-valued parameters, now we train the BNN, but with binarized inputs. For instance, instead of real values between $-1$ and $+1$ in the bipolar case, we take their sign as the bipolar binary features. As for the $0/1$ binaries, we simply round the pixel intensity. Fixed-point inputs are already binarized. Now we train the new BNN with the noisy backpropagation technique as described in \ref{sec:bnn_bp}. The second row of Table \ref{tab:mnist} shows the BNN results. We see that the bitwise networks perform well with very small additional errors. Note that the performance of the original real-valued dropout network with similar network topology (logistic units without max-norm constraint) is $1.35\%$.

\section{Conclusion}
In this work we propose a bitwise version of artificial neural networks, where all the inputs, weights, biases, hidden units, and outputs can be represented with single bits and operated on using simple bitwise logic. Such a network is very computationally efficient and can be valuable for resource-constrained situations, particularly in cases where floating-point / fixed-point variables and operations are prohibitively expensive. In the future we plan to investigate a bitwise version of convolutive neural networks, where efficient computing is more desirable.

\bibliography{mjkim.bib}
\bibliographystyle{icml2015}

\end{document}